\documentclass[11pt,a4paper]{article}
\usepackage{times,latexsym}
\usepackage{url}
\usepackage[T1]{fontenc}
\usepackage{authblk} 
\usepackage{bbm}
\makeatletter
\newcommand\email[2][]%
   {\newaffiltrue\let\AB@blk@and\AB@pand
      \if\relax#1\relax\def\AB@note{\AB@thenote}\else\def\AB@note{\relax}%
        \setcounter{Maxaffil}{0}\fi
      \begingroup
        \let\protect\@unexpandable@protect
        \def\thanks{\protect\thanks}\def\footnote{\protect\footnote}%
        \@temptokena=\expandafter{\AB@authors}%
        {\def\\{\protect\\\protect\Affilfont}\xdef\AB@temp{#2}}%
         \xdef\AB@authors{\the\@temptokena\AB@las\AB@au@str
         \protect\\[\affilsep]\protect\Affilfont\AB@temp}%
         \gdef\AB@las{}\gdef\AB@au@str{}%
        {\def\\{, \ignorespaces}\xdef\AB@temp{#2}}%
        \@temptokena=\expandafter{\AB@affillist}%
        \xdef\AB@affillist{\the\@temptokena \AB@affilsep
          \AB@affilnote{}\protect\Affilfont\AB@temp}%
      \endgroup
       \let\AB@affilsep\AB@affilsepx
}
\makeatother
\usepackage{graphicx}
\usepackage{framed}

\usepackage[acceptedWithA]{tacl2018v2}

\usepackage{amssymb}
\usepackage{enumitem}
\usepackage{comment}

\usepackage{xspace,mfirstuc,tabulary}

\newcommand\commentout[1]{}

\newif\iftaclinstructions
\taclinstructionsfalse 
\iftaclinstructions
\renewcommand{\confidential}{}
\renewcommand{\anonsubtext}{(No author info supplied here, for consistency with
TACL-submission anonymization requirements)}
\newcommand{\instr}
\fi

\newcommand{\answer}[1]{{\color{blue} #1}}

\iftaclpubformat 

\else

\fi

\usepackage{amsmath}
\usepackage{multirow}
\usepackage{booktabs}
\usepackage{graphicx}
\usepackage[normalem]{ulem}

\newif\ifcomments
\commentstrue
\ifcomments
    \providecommand\mg[1]{\textcolor{purple}{MG: #1}}
    \providecommand{\daniel}[1]{{\small \color{red} [DK: #1]}}
    \providecommand{\tushar}[1]{{\small \color{orange} [TK: #1]}}
    \providecommand{\jb}[1]{{\small \color{blue} [JB: #1]}}
    \providecommand{\es}[1]{{\small \color{teal} [ES: #1]}}
    
\else
    \providecommand{\mg}[1]{}
    \providecommand{\daniel}[1]{}
    \providecommand{\tushar}[1]{}
    \providecommand{\jb}[1]{}
    \providecommand{\es}[1]{}  
    \fi
\newcommand\nl[1]{{\it``#1''}}

\newcommand\strategyqa{\textsc{StrategyQA}}
\newcommand\ckptzero{\textsc{Ptd}}
\newcommand\fntd{\textsc{Fntd}}

\title{\emph{Did Aristotle Use a Laptop?} \\ 
A Question Answering Benchmark with Implicit Reasoning Strategies
}

\author[1,2]{\bf Mor Geva}
\author[2]{\bf Daniel Khashabi}
\author[1]{\bf Elad Segal}
\author[2]{\bf Tushar Khot}
\author[3]{\\\bf Dan Roth}
\author[1,2]{\bf Jonathan Berant}

{
\makeatletter
\renewcommand\AB@affilsepx{\quad \protect\Affilfont}
\makeatother
\affil[1]{Tel Aviv University}
\affil[2]{Allen Institute for AI}
\affil[3]{University of Pennsylvania}
}

\email{}

\email{\normalsize \texttt{morgeva@mail.tau.ac.il}, \quad \texttt{$\{$danielk,tushark$\}$@allenai.org},}
\email{\normalsize \texttt{elad.segal@gmail.com}, \quad \texttt{danroth@seas.upenn.edu}}
\email{\normalsize \texttt{joberant@cs.tau.ac.il}}

\date{}

\begin{document}
\maketitle

\begin{abstract}
A key limitation in current datasets for \emph{multi-hop reasoning} is that the required steps for answering the question are mentioned in it \emph{explicitly}.
In this work, we introduce \strategyqa{}, a question answering (QA) benchmark where the required reasoning steps are \emph{implicit} in the question, and should be inferred using a \emph{strategy}.
A fundamental challenge in this setup is how to elicit such creative questions from crowdsourcing workers, while covering a broad range of potential strategies. We propose a data collection procedure that combines term-based priming to inspire annotators, careful control over the annotator population, and adversarial filtering for eliminating reasoning shortcuts.
Moreover, we annotate each question with (1) a decomposition into reasoning steps for answering it, and 
(2) Wikipedia paragraphs that contain the answers to each step.
Overall, \strategyqa{} includes 2,780 examples, each consisting of a strategy question, its decomposition, and evidence paragraphs.
Analysis shows that questions in \strategyqa{} are short, topic-diverse, and cover a wide range of strategies.
Empirically, we show that humans perform well ($87$\%) on this task, while our best baseline reaches an accuracy of $\sim 66\%$. 

\end{abstract}

\section{Introduction}
\label{sec:introduction}

Developing models that successfully reason over multiple parts of their input has attracted substantial attention recently, leading to the creation of many multi-step reasoning Question Answering (QA) benchmarks
~\cite{welbl2018constructing,talmor2018web,khashabi2018looking,yang2018hotpotqa,dua2019drop,suhr2019corpus}.

\begin{figure}
    \centering
    \includegraphics[scale=0.72,trim=0cm 0.5cm 0cm 0cm]{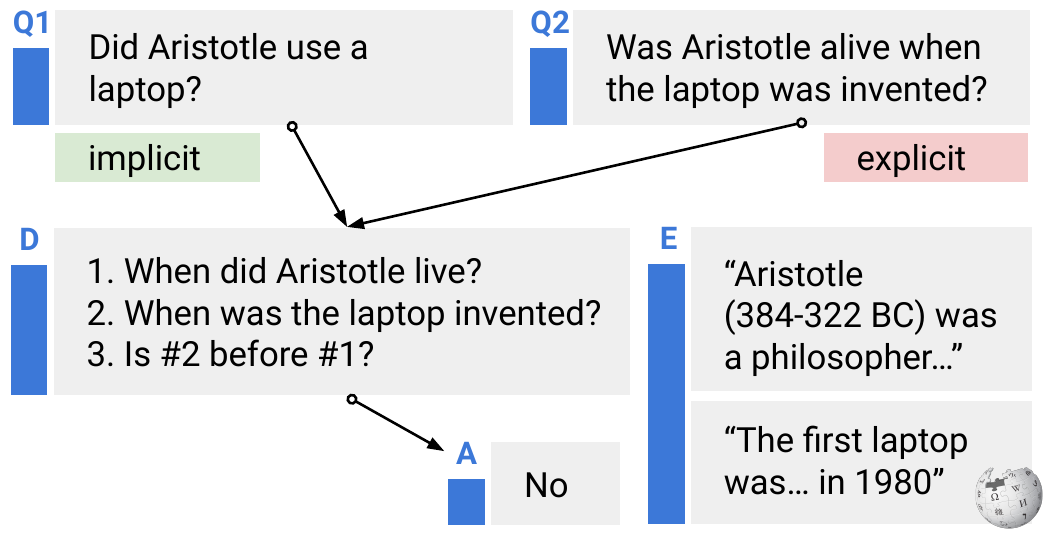}
    \caption{
        Questions in \strategyqa{} (Q1) require \emph{implicit} decomposition into reasoning steps (D), for which we annotate supporting evidence from Wikipedia (E). This is in contrast to multi-step questions that \emph{explicitly} specify the reasoning process (Q2). 
    }
    \label{figure:intro}
\end{figure}

Commonly, the language of questions in such benchmarks \emph{explicitly} describes the process for deriving the answer. For instance (Figure~\ref{figure:intro}, Q2), the question \nl{Was Aristotle alive when the laptop was invented?} explicitly specifies the required reasoning steps. However, in real-life questions, reasoning is often \emph{implicit}. For example, the question \nl{Did Aristotle use a laptop?} (Q1) can be answered using the same steps, but the model must infer the \emph{strategy} for answering the question -- \emph{temporal} comparison, in this case.

Answering implicit questions poses several challenges compared to answering their explicit counterparts. 
First, retrieving the context is difficult as there is little overlap between the question and its context (Figure~\ref{figure:intro}, Q1 and `E').  Moreover, questions tend to be short,  lowering the possibility of the model exploiting shortcuts in the language of the question.
In this work, we introduce \strategyqa{}, a boolean QA benchmark focusing on implicit multi-hop reasoning for \emph{strategy questions}, where a \emph{strategy} is the ability to infer from a question its atomic sub-questions. 
In contrast to previous benchmarks \cite{khot2020qasc, yang2018hotpotqa}, questions in \strategyqa{} are not limited to predefined decomposition patterns and cover a wide range of strategies that humans apply when answering questions.

Eliciting strategy questions using crowdsourcing is non-trivial.
First, authoring such questions requires \emph{creativity}. Past work often collected multi-hop questions by showing workers an entire context, which led to limited creativity and high lexical overlap between questions and contexts and consequently to reasoning shortcuts \cite{khot2020qasc, yang2018hotpotqa}. An alternative approach, applied in Natural Questions \cite{kwiatkowski2019natural} and MS-MARCO \cite{nguyen2016ms}, overcomes this by collecting real user questions. However, can we elicit creative questions independently of the context and without access to users?

Second, an important property in \strategyqa{} is that questions entail \emph{diverse} strategies. While the example in Figure~\ref{figure:intro} necessitates temporal reasoning, there are many possible strategies for answering questions (Table~\ref{table:example_strategy_questions}). We want a benchmark that exposes a broad range of strategies. But crowdsourcing workers often use repetitive patterns, which may limit question diversity.

To overcome these difficulties, we use the following techniques in our pipeline for eliciting strategy questions: (a) we prime crowd workers with random Wikipedia terms that serve as a minimal context to inspire their imagination and increase their creativity; (b) we use a large set of annotators to increase question diversity, limiting the number of questions a single annotator can write; and (c) we continuously train adversarial models during data collection, slowly increasing the difficulty in question writing and preventing recurring patterns~\cite{bartolo2020beat}.

Beyond the questions, as part of \strategyqa{}, we annotated: (a) \emph{question decompositions}: 
a sequence of steps sufficient for answering the question
(`D' in Figure~\ref{figure:intro}), and (b) \emph{evidence} paragraphs: Wikipedia paragraphs that contain the answer to \emph{each} decomposition step (`E' in Figure~\ref{figure:intro}). \strategyqa{} is the first QA dataset to provide decompositions and evidence annotations for each individual step of the reasoning process.

Our analysis shows that \strategyqa{} necessitates reasoning on a wide variety of knowledge domains (physics, geography, etc.) and logical operations (e.g. number comparison). Moreover, experiments show that \strategyqa{} poses a combined challenge of retrieval and QA, and while humans perform well on these questions, even strong systems struggle to answer them.

In summary, the contributions of this work are: 
\begin{enumerate}[noitemsep,nolistsep]
    \item Defining \emph{strategy} questions -- a class of question requiring \emph{implicit} multi-step reasoning.
    \item \strategyqa{}, the first benchmark for implicit multi-step QA, that covers a diverse set of reasoning skills. \strategyqa{} consists of 2,780 questions, annotated with their decomposition and per-step evidence.
    \item A novel annotation pipeline designed to elicit quality strategy questions, with minimal context for priming workers.
\end{enumerate}
The dataset and codebase are publicly available at \url{https://allenai.org/data/strategyqa}.

\section{Strategy Questions}
\label{sec:strategy_questions}

\subsection{Desiderata}
\label{sec:strategy_questions_desiderata}

\begin{table*}[t]
    \centering
    \footnotesize
    {
    \begin{tabular}{p{7cm}|p{8cm}}
         Question & Implicit facts \\ \toprule
         Can one spot helium? (\answer{No}) & Helium is a gas, Helium is odorless, Helium is tasteless, Helium has no color \\ \midrule
         Would Hades and Osiris hypothetically compete for real estate in the Underworld? (\answer{Yes}) & Hades was the Greek god of death and the Underworld. Osiris was the Egyptian god of the Underworld. \\ \midrule
        Would a monocle be appropriate for a cyclop? (\answer{Yes}) & Cyclops have one eye. A monocle helps one eye at a time. \\ \midrule
        Should a finished website have lorem ipsum paragraphs? (\answer{No}) & Lorem Ipsum paragraphs are meant to be temporary. Web designers always remove lorem ipsum paragraphs before launch. \\ \midrule
        Is it normal to find parsley in multiple sections of the grocery store? (\answer{Yes}) & Parsley is available in both fresh and dry forms. Fresh parsley must be kept cool. Dry parsley is a shelf stable product. \\ \hline
    \end{tabular}
    }
    \caption{Example strategy questions and the implicit facts needed for answering them. }
    \label{table:example_strategy_questions}
\end{table*}

\begin{table*}[t]
    \centering
    \footnotesize
    {
    \begin{tabular}{p{4.4cm}|c|c|p{9.0cm}}
         Question & MS & IM & Explanation\\ \toprule
         Was Barack Obama born in the United States?  (\answer{Yes}) & & & The question explicitly states the required information for the answer -- the birth place of Barack Obama. The answer is likely to be found in a single text fragment in Wikipedia. \\ \hline
         Do cars use drinking water to power their engine? (\answer{No}) & & & The question explicitly states the required information for the answer -- the liquid used to power car engines. The answer is likely to be found in a single text fragment in Wikipedia.  \\ \hline
         Are sharks faster than crabs? (\answer{Yes}) & \checkmark & & The question explicitly states the required reasoning steps: 1) How fast are sharks? 2) How fast are crabs? 3) Is \#1 faster than \#2? \\ \hline
        Was Tom Cruise married to the female star of Inland Empire? (\answer{No}) & \checkmark & & The question explicitly states the required reasoning steps: 1) Who is the female star of Inland Empire? 2) Was Tom Cruise married to \#2? \\  \hline
         Are more watermelons grown in Texas than in Antarctica? (\answer{Yes}) & \checkmark & \checkmark & The answer can be derived through geographical/botanical reasoning that the climate in Antarctica does not support growth of watermelons. \\ \hline
         Would someone with a nosebleed benefit from Coca? (\answer{Yes}) & \checkmark & \checkmark & The answer can be derived through biological reasoning that Coca constricts blood vessels, and therefore, serves to stop bleeding. \\ \hline
    \end{tabular}
    }
    \caption{Example questions demonstrating the multi-step (MS) and implicit (IM) properties of strategy questions.}
    \label{table:example_question_properties}
\end{table*}

We define strategy questions by characterizing their desired properties. 
Some properties, such as whether the question is answerable, also depend on the context used for answering the question. 
In this work, we assume this context is a corpus of documents, specifically, Wikipedia, which we assume provides correct content.

\paragraph{Multi-step} 
Strategy questions are multi-step questions, that is, they comprise a sequence of \emph{single-step questions}. A single-step question is either (a) a question that can be answered from a short text fragment in the corpus (e.g. steps 1 and 2 in Figure~\ref{figure:intro}), or (b) a logical operation over answers from previous steps (e.g. step 3 in Figure~\ref{figure:intro}). A strategy question should have at least two steps for deriving the answer. Example multi- and single- step questions are provided in Table~\ref{table:example_question_properties}.
We define the reasoning process structure in \S\ref{sec:strategy_questions_supervision}.

\paragraph{Feasible}
Questions should be answerable from paragraphs in the corpus. Specifically, for each reasoning step in the sequence, there should be sufficient evidence from the corpus to answer the question.
For example, the answer to the question \nl{Would a monocle be appropriate for a cyclop?} can be derived from paragraphs stating that cyclops have one eye and that a monocle is used by one eye at the time. This information is found in our corpus, Wikipedia, and thus the question is feasible.
In contrast, the question \nl{Does Justin Beiber own a Zune?} is \emph{not} feasible, because answering it requires going through Beiber's belongings, and this information is unlikely to be found in Wikipedia.

\paragraph{Implicit} A key property distinguishing strategy questions from prior multi-hop questions is their implicit nature.
In explicit questions, each step in the reasoning process can be inferred from the \emph{language} of the question directly. For example, in Figure~\ref{figure:intro}, the first two questions are explicitly stated, one in the main clause and one in the adverbial clause. Conversely, reasoning steps in strategy questions require going beyond the language of the question. Due to language variability, a precise definition of implicit questions based on lexical overlap is elusive, but a good rule-of-thumb is the following: if the question decomposition can be written with a vocabulary limited to words from the questions, their inflections, and function words, then it is an explicit question. If new content words must be introduced to describe the reasoning process, the question is implicit.
Examples for implicit and explicit questions are in Table~\ref{table:example_question_properties}.

\commentout{
\paragraph{Feasible}
Questions should be answerable from paragraphs in the corpus. Specifically, there should exist a reasoning process for which there is sufficient evidence from the corpus to answer the question.
For example, to answer the question \nl{Would a monocle be appropriate for a cyclop?} could be derived from paragraphs stating that cyclops have one eye and that a monocle helps one eye at the time. Paragraphs providing this information can be found in Wikipedia, thus, the question is feasible.
In contrast, the question \nl{Does Justin Beiber own a Zune?} is \emph{not} feasible, because the only way to answer it would be to go through the list of Beiber's belongings, which is unlikely to appear in Wikipedia.

}
\paragraph{Definite} 
A type of questions we wish to avoid are \emph{non-definitive} questions, such as \nl{Are hamburgers considered a sandwich?} and \nl{Does chocolate taste better than vanilla?} for which there is no clear answer.
We would like to collect questions where the answer is definitive or, at least, very likely, based on the corpus.
E.g., consider the question \nl{Does wood conduct electricity?}. Although it is possible that a damp wood will conduct electricity, the answer is generally \emph{no}.

To summarize, strategy questions are multi-step questions with implicit reasoning (a strategy) and a definitive answer 
that can be reached given a corpus.
We limit ourselves to Boolean
yes/no questions,
which limits the output space, but lets us focus on the complexity of the questions, which is the key contribution. 
Example strategy questions are in Table~\ref{table:example_strategy_questions}, and examples that demonstrate the mentioned properties are in Table~\ref{table:example_question_properties}.
Next (\S\ref{sec:strategy_questions_supervision}), we describe additional structures annotated during data collection.

\subsection{Decomposing Strategy Questions}
\label{sec:strategy_questions_supervision}


Strategy questions involve complex reasoning that leads to a yes/no answer.
To guide and evaluate the QA process, we annotate every example with a description of the expected reasoning process. 

Prior work used \emph{rationales} or \emph{supporting facts}, i.e., text snippets extracted from the context \cite{deyoung2020eraser,yang2018hotpotqa, kwiatkowski2019natural,khot2020qasc} as evidence for an answer. 
However, reasoning can rely on elements that are not explicitly expressed in the context.
Moreover, answering a question based on relevant context does not imply that the model performs reasoning properly \cite{jiang2019avoiding}. 

Inspired by recent work \cite{wolfson2020break}, we associate every question-answer pair with a \emph{strategy question decomposition}.
A decomposition of a question $q$ is a sequence of $n$ steps $\langle s^{(1)}, s^{(2)}, ..., s^{(n)} \rangle$ required for computing the answer to $q$. 
Each step $s^{(i)}$ corresponds to a single-step question and may include special \emph{references}, which are placeholders referring to the result of a previous step $s^{(j)}$. The last decomposition step (i.e. $s^{(n)}$) returns the final answer to the question. Table~\ref{table:decomposition_examples} shows decomposition examples. 

\newcite{wolfson2020break} targeted explicit multi-step questions (first row in Table~\ref{table:decomposition_examples}), where the decomposition is restricted to a small vocabulary derived almost entirely from the original question.
Conversely, decomposing strategy questions requires using implicit knowledge, and thus decompositions can include any token that is needed for describing the implicit reasoning (rows 2-4 in Table~\ref{table:decomposition_examples}). This makes the decomposition task significantly harder for strategy questions.

In this work, we distinguish between two types of required actions for executing a step. \textit{Retrieval}: a step that requires retrieval from the corpus, and  \textit{operation}, a logical function over answers to previous steps. 
In the second row of Table~\ref{table:decomposition_examples}, the first two steps are retrieval steps, and the last step is an operation. 
A decomposition step can require both retrieval and an operation (see last row in Table~\ref{table:decomposition_examples}).


\begin{table}[t]
    \centering
    \footnotesize
    {
    \begin{tabular}{p{1.9cm}|p{4.8cm}}
        Question & Decomposition \\ \toprule
         \multirow{3}{=}{Did the Battle of Peleliu or the Seven Days Battles last longer?} & (1) How long did \emph{the Battle of Peleliu last}? \\
         & (2) How long did \emph{the Seven Days Battle last}? \\
         & (3) Which is \emph{longer} of \#1 , \#2? \\ \midrule \midrule 
         \multirow{3}{=}{Can the President of Mexico vote in New Mexico primaries?} & (1) What is \textbf{the citizenship requirement} for \emph{voting in} \emph{New Mexico}? \\
         & (2) What is \textbf{the citizenship requirement} of any \emph{President of Mexico}? \\
         & (3) Is \#2 the same as \#1?  \\ \midrule
         \multirow{4}{=}{Can a microwave melt a Toyota Prius battery?} & (1) What \textbf{kind of} \emph{battery} does a \emph{Toyota Prius} use? \\
         & (2) What \textbf{type of material} is \#1 made out of? \\
         & (3) What is the \textbf{melting point} of \#2?  \\
         & (4) Can a \emph{microwave's} \textbf{temperature} reach at least \#3?    \\ \midrule
        \multirow{3}{=}{Would it be common to find a penguin in Miami?} & (1) Where is a typical \emph{penguin's} \textbf{natural habitat}? \\
         & (2) What \textbf{conditions} make \#1 suitable for \emph{penguins}? \\
         & (3) Are all of \#2 present in \emph{Miami}?
    \end{tabular}
    }
    \caption{Explicit (row 1) and strategy (rows 2-4) question decompositions. We mark words that are explicit (italic) or implicit in the input (bold).
    }
    \label{table:decomposition_examples}
\end{table}

To verify that steps are valid single-step questions that can be answered using the corpus (Wikipedia), we collect
\textit{supporting evidence} for each retrieval step and annotate operation steps.
A supporting evidence is one or more paragraphs that provide an answer to the retrieval step. 

In summary, each example in our dataset contains a) a strategy question, b) the strategy question decomposition, and c) supporting evidence per decomposition step. 
Collecting strategy questions and their annotations is the main challenge of this work, and we turn to this next.

\section{Data Collection Pipeline}
\label{sec:data_collection}

Our goal is to establish a procedure
for collecting strategy questions and their annotations at scale. To this end, we build a multi-step crowdsourcing\footnote{We use Amazon Mechanical Turk as our framework.} pipeline designed for encouraging worker creativity, while preventing biases in the data.

\begin{figure*}
    \centering
    \includegraphics[scale=0.8]{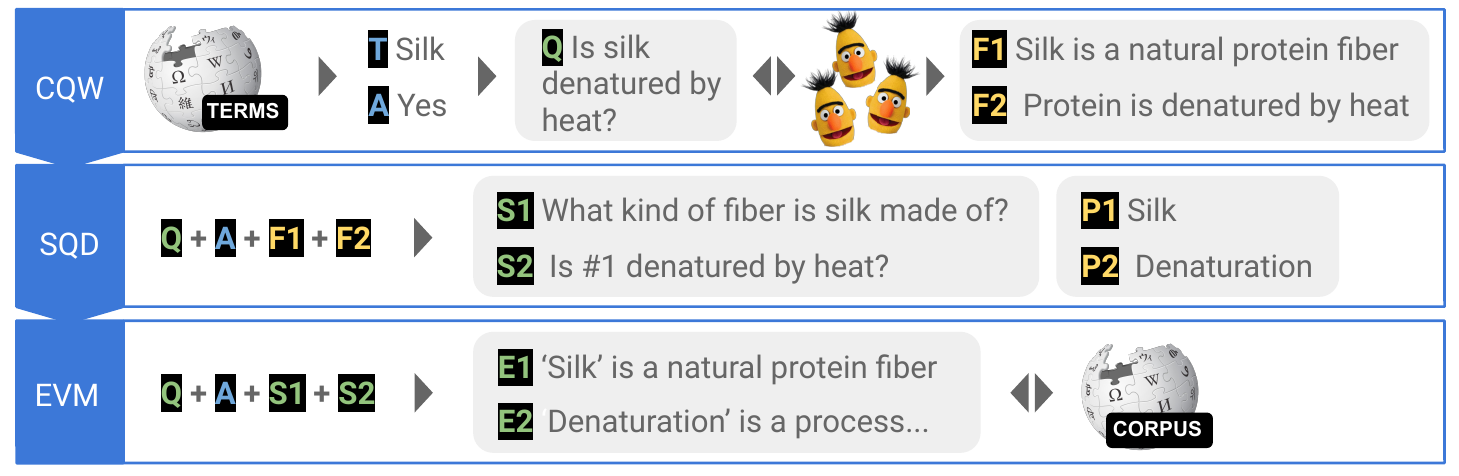}
    \vspace*{-2mm}
    \caption{Overview of the data collection pipeline. First (CQW, \S\ref{subsection:cqw}), a worker is presented with a term (T) and an expected answer (A) and writes a question (Q) and the facts (F1,F2) required to answer it. Next, the question is decomposed (SQD, \S\ref{subsection:sqd}) into steps (S1, S2) along with Wikipedia page titles (P1,P2) that the worker expects to find the answer in. Last (EVM, \S\ref{subsection:evm}), decomposition steps are matched with evidence from Wikipedia (E1, E2).
    }
    \label{figure:pipeline}
\end{figure*}

We break the data collection into three tasks: question writing (\S\ref{subsection:cqw}), question decomposition (\S\ref{subsection:sqd}), and evidence matching (\S\ref{subsection:evm}). In addition, we implement mechanisms for quality assurance (\S\ref{subsection:verification}).
An overview of the data collection pipeline is in Figure~\ref{figure:pipeline}. 


\subsection{Creative Question Writing (CQW)}
\label{subsection:cqw}

Generating natural language annotations  through crowdsourcing (e.g.,  question generation) is known to suffer from several shortcomings. First, when annotators generate many instances, they use recurring patterns that lead to biases in the data. \cite{gururangan2018annotation,geva2019modeling}.
Second, when language is generated conditioned on a long context, such as a paragraph, annotators use similar language \cite{kwiatkowski2019natural}, leading to high lexical overlap and hence, inadvertently, to an easier problem.    
Moreover, a unique property of our setup is that we wish to cover a \emph{broad and diverse} set of strategies. Thus, we must discourage repeated use of the same strategy. 

We tackle these challenges on multiple fronts. First, rather than using a long paragraph as context,  we prime workers to write questions given single terms from Wikipedia, reducing the overlap with the context to a minimum.
Second, to encourage diversity, we control the population of annotators, making sure a large number of annotators contribute to the dataset. Third, we use \emph{model-in-the-loop} adversarial annotations \cite{dua2019drop,khot2020qasc,bartolo2020beat} to filter our questions, and only accept questions that fool our models. While some model-in-the-loop approaches use fixed pre-trained models to eliminate ``easy'' questions, we continuously update the models during data collection to combat the use of repeated patterns or strategies.


We now provide a description of the task, and elaborate on these methods (Figure~\ref{figure:pipeline}, upper row). 

\paragraph{Task description} Given a term (e.g., \emph{silk}), a description of the term, and an expected answer (yes or no), the task is to write a strategy question about the term with the expected answer, and the facts required to answer the question.

\paragraph{Priming with Wikipedia terms}
Writing strategy questions from scratch is difficult. To inspire worker creativity, we ask to write questions about terms they are familiar with or can easily understand.
The terms are titles of ``popular''\footnote{We filter pages based on the number of contributors and the number of backward links from other pages.} Wikipedia pages.
We provide workers only with a short description of the given term.
Then, workers use their background knowledge and web search skills to form a strategy question.

\paragraph{Controlling the answer distribution} We ask workers to write questions where the answer is set to be `yes' or `no'. To balance the answer distribution, the expected answer is dynamically sampled inversely proportional to the ratio of `yes' and `no' questions collected until that point.

\paragraph{Model-in-the-loop filtering}
To ensure questions are challenging and reduce recurring language and reasoning patterns, questions are only accepted when verified by two sets of online \emph{solvers}. We deploy a set of 5 pre-trained models (termed \ckptzero{}) that check if the question is too easy. If at least 4 out of 5 answer the question correctly, it is rejected. 
Second, we use a set of 3 models (called \fntd{}) that are continuously fine-tuned on our collected data and are meant to detect biases in the current question set. 
A question is rejected if all 3 solvers answer it correctly. 
The solvers are \textsc{RoBERTa} \cite{liu2019roberta} models fine-tuned on different auxiliary datasets; details in \S\ref{subsec:baselines}.

\paragraph{Auxiliary sub-task} We ask workers to provide the facts required to answer the question they have written,  
for several reasons: 1) it helps workers frame the question writing task and describe the reasoning process they have in mind, 2) it helps reviewing their work, and 3) it provides useful information for the decomposition step (\S\ref{subsection:sqd}).



\subsection{Strategy Question Decomposition (SQD)}
\label{subsection:sqd}

Once a question and the corresponding facts are written, we generate the strategy question decomposition (Figure~\ref{figure:pipeline}, middle row). 
We annotate decompositions \emph{before} matching evidence in order to avoid biases stemming from seeing the context.


The decomposition strategy for a question is not always obvious, which can lead to undesirable explicit decompositions. For example, a possible explicit decomposition for Q1 (Figure~\ref{figure:intro}) might be (1) \textit{What items did Aristotle use?} (2) \textit{Is laptop in \#1?}; but the first step is not feasible.
To guide the decomposition, we provide workers with the facts written in the CQW task to show the strategy of the question author. Evidently, there can be many valid strategies and the same strategy can be phrased in multiple ways -- the facts only serve as a soft guidance. 

\paragraph{Task description} Given a strategy question, a yes/no answer, and a set of facts, the task is to write the steps needed to answer the question.

\paragraph{Auxiliary sub-task} 
We observe that in some cases, annotators write explicit decompositions, which often lead to infeasible steps that cannot be answered from the corpus. 
To help workers avoid explicit decompositions, we ask them to specify, for each decomposition step, a Wikipedia page they expect to find the answer in. This encourages workers to write decomposition steps for which it is possible to find answers in Wikipedia, and leads to feasible strategy decompositions, with only a small overhead (the workers are not required to read the proposed Wikipedia page).


\subsection{Evidence Matching (EVM)}
\label{subsection:evm}
We now have a question and its decomposition. To ground them in context, we add a third task of evidence matching (Figure~\ref{figure:pipeline}, bottom row).

\paragraph{Task description} Given a question and its decomposition (a list of single-step questions), the task is to find evidence paragraphs on Wikipedia for each retrieval step. Operation steps that do not require retrieval (\S\ref{sec:strategy_questions_supervision}) are marked as  \emph{operation}.

\paragraph{Controlling the matched context} Workers search for evidence on Wikipedia. 
We index Wikipedia\footnote{We use the Wikipedia Cirrus dump from 11/05/2020. 
} and provide a search interface where workers can drag-and-drop paragraphs from the results shown on the search interface. 
This guarantees that annotators choose paragraphs we included in our index, at a pre-determined paragraph-level granularity. 


\subsection{Data Verification Mechanisms}
\label{subsection:verification}


\paragraph{Task qualifications} For each task, we hold qualifications that test understanding of the task, and manually review several examples. Workers who follow the requirements are granted access to our tasks. Our qualifications are open to workers from English speaking countries who have high reputation scores. 
Additionally, the authors regularly review annotations to give feedback and prevent noisy annotations.

\paragraph{Real-time automatic checks} 
For CQW, we use heuristics to check question validity, e.g., whether it ends with a question mark, and that it doesn't use language that characterizes explicit multi-hop questions (for instance, having multiple verbs). 
For SQD, we check that the decomposition structure forms a directed acyclic graph, i.e. (i) each decomposition step is referenced by (at least) one of the following steps, such that all steps are reachable from the last step; and (ii) steps don't form a cycle. 
In the EVM task, a warning message is shown when the worker marks an intermediate step as an operation (an unlikely scenario).

\paragraph{Inter-task feedback} At each step of the pipeline, we collect feedback about previous steps.
To verify results from the CQW task, we ask workers to indicate whether the given answer is incorrect (in the SQD, EVM tasks), or if the question is not definitive (in the SQD task) (\S\ref{sec:strategy_questions_desiderata}).
Similarly, to identify non-feasible questions or decompositions, we ask workers to indicate if there is no evidence for a decomposition step (in the EVM task).

\paragraph{Evidence verification task} 
After the EVM step, each example comprises a question, its answer, decomposition and supporting evidence.
To verify that a question can be answered by executing the decomposition steps against the matched evidence paragraphs, we construct an additional evidence verification task (EVV). In this task, workers are given a question, its decomposition and matched paragraphs, and are asked to answer the question in each decomposition step purely based on the provided paragraphs. Running EVV on a subset of examples during data collection, helps identify issues in the pipeline and in worker performance. 



\section{The \strategyqa{} Dataset}
\label{sec:strategyqa}

\begin{table}[t]
    \centering
    \footnotesize
    {
    \begin{tabular}{p{4.6cm}|c|c}
         & Train & Test \\ \toprule
        \# of questions  & 2290 & 490 \\
        \% ``yes'' questions & 46.8\% & 46.1\% \\  
        \# of unique terms  & 1333 & 442 \\
        \# of unique decomposition steps  & 6050 & 1347 \\ 
        \# of unique evidence paragraphs  & 9251 & 2136 \\
        \# of occurrences of the top trigram   & 31 & 5 \\ \hline
        \# of question writers  &  23 & 6 \\
        \# of filtered questions  & 2821 & 484 \\ \hline
        Avg. question length (words)  & 9.6 & 9.8 \\
        Avg. decomposition length (steps) & 2.93 & 2.92 \\
        Avg. \# of paragraphs per question  & 2.33 & 2.29 \\
    \end{tabular}
    }
    \caption{\strategyqa{} statistics. Filtered questions were rejected by the solvers 
    (\S\ref{subsection:cqw}). The train and test sets of question writers are disjoint. 
    The ``top trigram'' is the most common trigram.
    }
    \label{table:dataset_statistics}
\end{table}

We run our pipeline on 1,799 Wikipedia terms, allowing a maximum of 5 questions per term. We update our online fine-tuned solvers (\textsc{Fntd}) every 1K questions. Every question is decomposed once, and evidence is matched for each decomposition by 3 different workers. The cost of annotating a full example is \$4.


To encourage diversity in strategies used in the questions, we recruited new workers throughout data collection. 
Moreover, periodic updates of the online solvers prevent workers from exploiting shortcuts, since the solvers adapt to the training distribution.
Overall, there were 29 question writers, 19 decomposers, and 54 evidence matchers participating in the data collection.

We collected 2,835 questions, out of which 55 were marked as having an incorrect answer during SQD (\S\ref{subsection:sqd}).
This results in a collection of 2,780 verified strategy questions, for which we create an annotator-based data split \cite{geva2019modeling}. 
We now describe the dataset statistics (\S\ref{subsec:dataset_statistics}), analyze the quality of the examples, (\S\ref{subsec:data_quality}) and explore the reasoning skills in \strategyqa{} (\S\ref{subsec:data_diversity}).

\subsection{Dataset Statistics}
\label{subsec:dataset_statistics}

We observe (Table~\ref{table:dataset_statistics}) that the answer distribution is roughly balanced (yes/no). Moreover, questions are short ($<10$ words), and the most common trigram occurs in roughly $1\%$ of the examples. This indicates that the language of the questions is both simple and diverse. For comparison, the average question length in the multi-hop datasets \textsc{HotpotQA} \cite{yang2018hotpotqa} and \textsc{ComplexWebQuestions} \cite{talmor2018web} is $13.7$ words and $15.8$ words, respectively. Likewise, the top trigram in these datasets occurs in 9.2\% and 4.8\% of their examples, respectively.

More than half of the generated questions are filtered by our solvers, pointing to the difficulty of generating good strategy questions. We release all 3,305 filtered questions as well.

\begin{figure}[t]
    \centering
    \includegraphics[scale=0.42, trim=0cm 0cm 0cm 0cm]{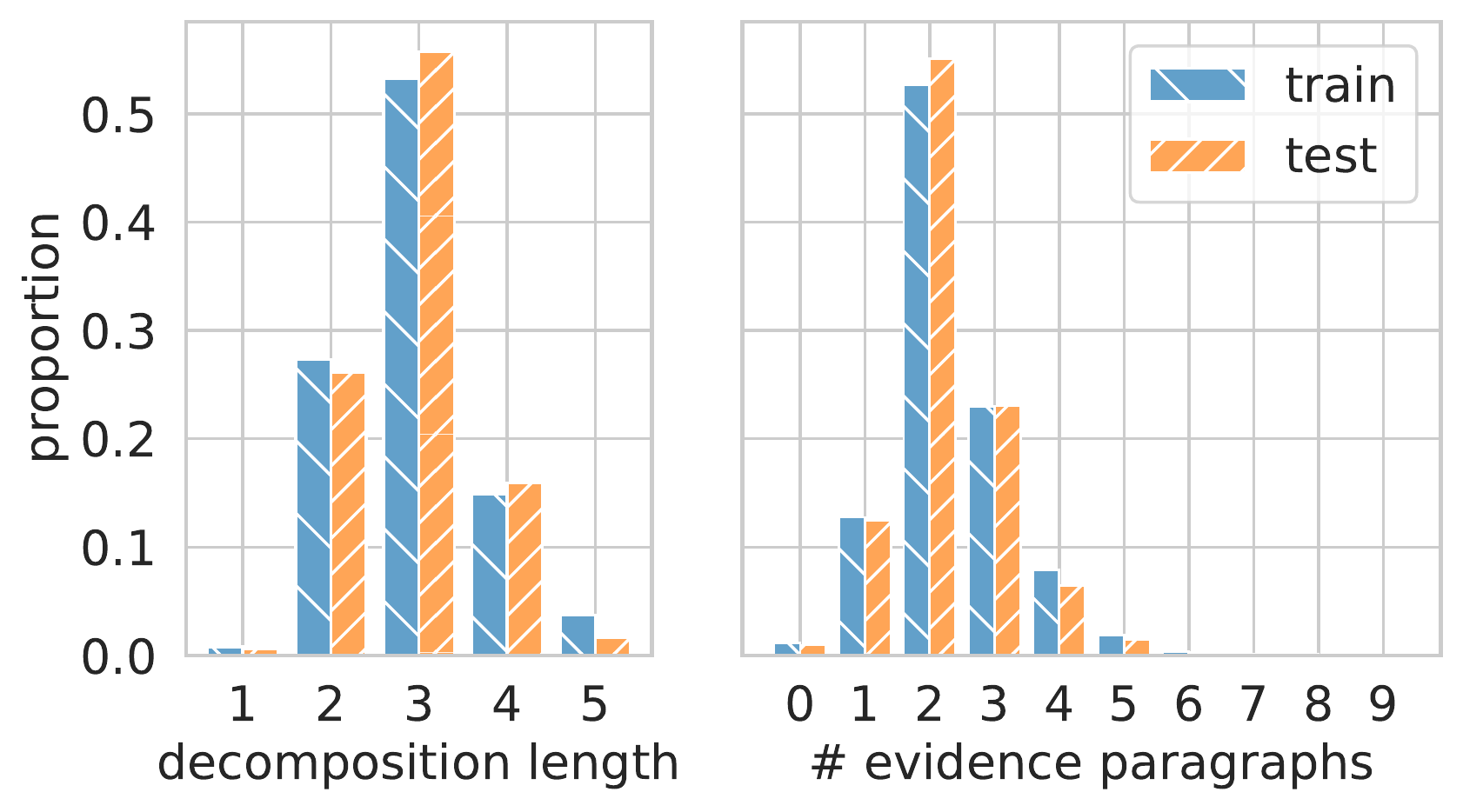}
    \vspace*{-2mm}
    \caption{
        The distributions of decomposition length (left) and the number of evidence paragraphs (right).
        The majority of the questions in \strategyqa{} require a reasoning process comprised of $\geq 3$ steps, of which about 2 steps involve retrieving external knowledge.
    }
    \label{figure:reasoning_process_complexity}
\end{figure}

To characterize the \emph{reasoning complexity} required to answer questions in \strategyqa{}, we examine the decomposition length and the number of evidence paragraphs. Figure~\ref{figure:reasoning_process_complexity} and Table~\ref{table:dataset_statistics} (bottom) show the distributions of these properties
are centered around 3-step decompositions and 2 evidence paragraphs, but a considerable
portion of the dataset requires more steps and paragraphs.




\begin{table}[t]
    \centering
    \footnotesize
    \begin{tabular}{l|c|c|c}
        & multi-step & single-step &  \\ \hline
      implicit  & 81 & 1 & 82 \\ \hline
      explicit  & 14.5 & 3.5 & 18 \\ \hline
       & 95.5 & 4.5 & 100
    \end{tabular}
    \caption{Distribution over the implicit and multi-step properties (\S\ref{sec:strategy_questions}) in a sample of 100 \strategyqa{} questions, annotated by two experts (we average the expert decisions).
    Most questions are \emph{multi-step} and \emph{implicit}. Annotator agreement is substantial for both the implicit ($\kappa = 0.73$) and multi-step ($\kappa = 0.65$) properties. 
    }
    \label{table:quality_stats}
\end{table}

\subsection{Data Quality}
\label{subsec:data_quality}

\paragraph{Do questions in \strategyqa{} require multi-step implicit reasoning?}
To assess the quality of questions, we sampled 100 random examples from the training set, and had two experts (authors) independently annotate whether the questions satisfy the desired properties of strategy questions (\S\ref{sec:strategy_questions_desiderata}).
We find that most of the examples (81\%) are valid multi-step implicit questions, 82\% of questions are implicit, and 95.5\% are multi-step (Table~\ref{table:quality_stats}).

\paragraph{Do questions in \strategyqa{} have a definitive answer?}
We let experts review the answers to 100 random questions, allowing access to the Web. We then ask them to state for every question whether they agree or disagree with the provided answer. We find that the experts agree with the answer in 94\% of the cases, and disagree only in 2\%. For the remaining 4\%, either the question was ambiguous, or the annotators could not find a definite answer on the Web. Overall,
this suggests that questions in \strategyqa{} have clear answers.

\paragraph{What is the quality of the decompositions?}
We randomly sampled 100 decompositions and asked experts to judge their quality.
Experts judged if the decomposition is explicit or utilizes a strategy. We find that 83\% of the decompositions validly use a strategy to break down the question. The remaining 17\% decompositions are explicit, however, in 14\% of the cases the original question is already explicit. Second, experts checked if the phrasing of the decomposition is ``natural'', i.e., it reflects the decomposition of a person that does not already know the answer. We find that 89\% of the decompositions express a ``natural'' reasoning process, while 11\% may depend on the answer. Last, we asked experts to indicate any potential logical flaws in the decomposition, but no such cases occurred in the sample.

\commentout{
A decomposition should express an implicit yet natural reasoning process, one would employ to derive an answer to the question.
To verify the correctness and coherence of decompositions in \strategyqa{}, we sample 100 decompositions, and ask experts (authors) to label every decomposition with two labels: a) strategic/explicit for whether the decomposition uses a strategy or is an explicit breakdown of the question, b) natural/artificial for whether the decomposition sounds like a natural breakdown of the question \jb{this sounds highly subjective...}. In addition, we ask experts to indicate any potential logical flaws in the decomposition.

We find that 83\% of the decompositions validly use a strategy to break down the question. The rest 17\% decompositions are explicit, however, in 14\% of the cases the question is by itself explicit. Moreover, 89\% of the decompositions express a natural reasoning process, while 11\% assume additional knowledge which makes them sound artificial. Lastly, there were no decompositions with logical flaws, that for example, could lead to the opposite answer.
}

\paragraph{Would different annotators use the same decomposition strategy?}
We sample 50 examples, and let two different workers decompose the questions. 
Comparing the decomposition pairs, we find that a) for all pairs, the last step returns the same answer, b) in 44 out of 50 pairs, the decomposition pairs follow the same reasoning path , and c) in the other 6 pairs, the decompositions either follow a different reasoning process (5 pairs) or one of the decompositions is explicit (1 pair). 
This shows that different workers usually use the same strategy when decomposing questions. 


\paragraph{Is the evidence for strategy questions in Wikipedia?}
Another important property is whether questions in \strategyqa{} can be answered based on context from our corpus, Wikipedia, given that questions are written independently of the context. 
To measure evidence coverage, in the EVM task (\S\ref{subsection:evm}), we provide workers with a checkbox for every decomposition step, indicating whether only partial or no evidence could be found for that step. Recall that three different workers match evidence for each decomposition step.
We find that 88.3\% of the questions are fully covered: evidence was matched for each step by some worker.
Moreover, in 86.9\% of the questions, at least one worker found evidence  for \emph{all} steps. Last, in only 0.5\% of the examples, all three annotators could not match evidence for \emph{any} of the steps. 
This suggests that overall, Wikipedia is a good corpus for questions in \strategyqa{}, that were written independently of the context.

\paragraph{Do matched paragraphs provide evidence?}
We assess the quality of matched paragraphs by analyzing both example-level and step-level annotations.
First, we sample 217 decomposition steps with their corresponding paragraphs matched by one of the three workers. We let 3 different crowdworkers decide whether the paragraphs provide evidence for the answer to that step. We find that in 93\% of the cases, the majority vote is that the evidence is valid.\footnote{With moderate annotator agreement of $\kappa = 0.42$.}


Next, we analyze annotations of the verification task (\S\ref{subsection:verification}), where workers are asked to answer all decomposition steps based only on the matched paragraphs. We find that the workers could answer sub-questions and derive the correct answer in 82 out of 100 annotations. Moreover, in 6 questions indeed there was an error in evidence matching, but another worker that annotated the example was able to compensate for the error, leading to 88\% of the questions where evidence matching succeeds. In the last 12 cases indeed evidence is missing, and is possibly absent from Wikipedia.

Lastly, we let experts review the paragraphs matched by one of the three workers to all the decomposition steps of a question, for 100 random questions. We find that for 79 of the questions the matched paragraphs provide sufficient evidence for answering the question. 
For 12 of the 21 questions without sufficient evidence, the experts indicated they would expect to find evidence in Wikipedia, and the worker probably could not find it. For the remaining 9 questions, they estimated that evidence is probably absent from Wikipedia.

In conclusion, 93\% of the paragraphs matched at the step-level were found to be valid. Moreover, when considering single-worker annotations, $\sim$80\% of the questions are matched with paragraphs that provide sufficient evidence for all retrieval steps. This number increases to 88\% when aggregating the annotations of three workers.

\paragraph{Do different annotators match the same evidence paragraphs?}
To compare the evidence paragraphs matched by different workers, we check whether for a given decomposition step, the same paragraph IDs are retrieved by different annotators.
Given two non-empty sets of paragraph IDs $\mathcal{P}_1, \mathcal{P}_2$, annotated by two workers, we compute the Jaccard coefficient $J(\mathcal{P}_1, \mathcal{P}_2) = \frac{|\mathcal{P}_1 \cap \mathcal{P}_2|}{|\mathcal{P}_1 \cup \mathcal{P}_2|}$. In addition, we take the sets of corresponding Wikipedia page IDs $\mathcal{T}_1, \mathcal{T}_2$ for the matched paragraphs, and compute $J(\mathcal{T}_1, \mathcal{T}_2)$. 
Note that a score of 1 is given to two identical sets, while a score of 0 corresponds to sets that are disjoint.
The average similarity score is 0.43 for paragraphs and 0.69 for pages. This suggests that evidence for a decomposition step can be found in more than one paragraph in the same page, or in different pages.

\begin{table}[t]
    \centering
    \footnotesize
    \begin{tabular}{lp{4.3cm}c}
      Strategy & Example & \% \\ \toprule
      Physical & \nl{Can human nails carve a statue out of quartz?} & 13 \\ \hline
      Biological & \nl{Is a platypus immune from cholera?} & 11 \\ \hline
      Historical & \nl{Were mollusks an ingredient in the color purple?} & 10 \\ \hline
      Temporal & \nl{Did the 40th president of the United States forward lolcats to his friends?} & 10 \\ \hline
      Definition & \nl{Are quadrupeds represented on Chinese calendar?} & 8 \\ \hline
      Cultural & \nl{Would a compass attuned to Earth's magnetic field be a bad gift for a Christmas elf?} & 5 \\ \hline
      Religious & \nl{Was Hillary Clinton's deputy chief of staff in 2009 baptised?} & 5 \\ \hline
      Entertainment & \nl{Would Garfield enjoy a trip to Italy?} & 4 \\ \hline
      Sports & \nl{Can Larry King's ex-wives form a water polo team?} & 4 \\ \hline
    \end{tabular}
    \caption{Top strategies in \strategyqa{} and their frequency in a 100 example subset (accounting for 70\% of the analyzed examples).}
    \label{table:top_strategies}
\end{table}

\begin{figure}[t]
    \centering
    \includegraphics[scale=0.33, trim=0cm 1cm 0cm 1cm]{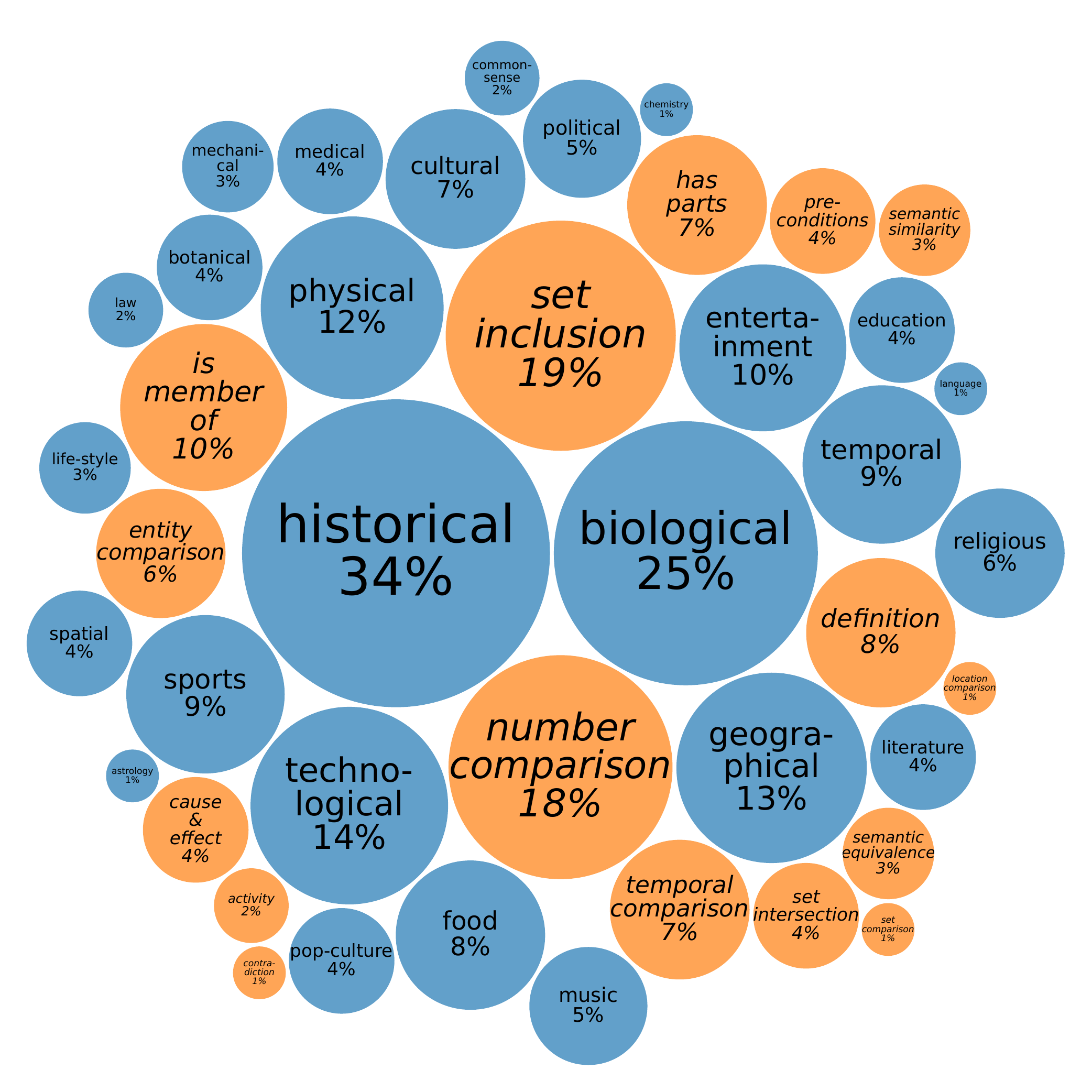}
    \caption{Reasoning skills in \strategyqa{}; each skill is associated with the proportion of examples it is required for. Domain-related and logical reasoning skills are marked in blue and orange (italic), respectively.}
    \label{figure:reasoning_skills}
\end{figure}

\begin{figure}[t]
    \centering
    \includegraphics[scale=0.4, trim=0cm 1cm 0cm 0cm]{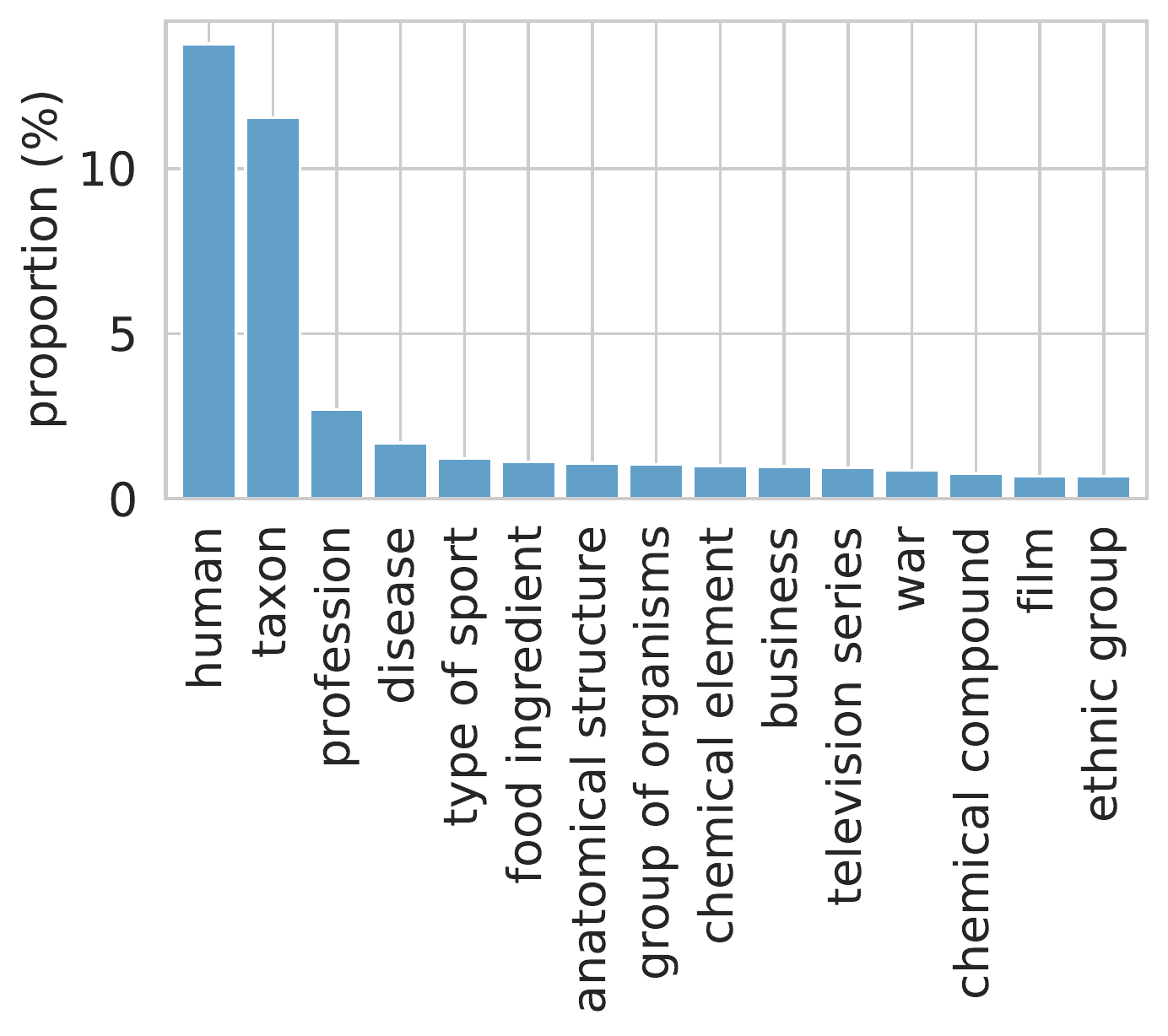}
    \caption{The top 15 categories of terms used to prime workers for question writing and their proportion.}
    \label{figure:term_categories}
\end{figure}

\subsection{Data Diversity}
\label{subsec:data_diversity}
We aim to generate creative and diverse questions. We now analyze diversity in terms of the required reasoning skills and question topic.

\paragraph{Reasoning skills}
To explore the required reasoning skills in \strategyqa{}, we sampled 100 examples and let two experts (authors) discuss and annotate each example with a) the type of strategy for decomposing the question, and b) the required reasoning and knowledge skills per decomposition step. We then aggregate similar labels (e.g. botanical $\rightarrow$ biological) and compute the proportion of examples each strategy/reasoning skill is required for (an example can have multiple strategy labels).

Table~\ref{table:top_strategies} demonstrates the top strategies, showing that \strategyqa{} contains a broad set of strategies.
Moreover,  diversity is apparent (Figure~\ref{figure:reasoning_skills}) in terms of both domain-related reasoning (e.g. biological and technological) and logical functions (e.g. set inclusion and ``is member of''). 
While the reasoning skills sampled from questions in \strategyqa{} do not necessarily reflect their prevalence in a ``natural'' distribution, we argue that promoting research on methods for inferring strategies is an important research direction.

\paragraph{Question topics}
As questions in \strategyqa{} were triggered by Wikipedia terms, we use the ``instance of'' Wikipedia property to characterize the topics of questions.\footnote{It is usually a 1-to-1 mapping from a term to a Wikipedia category. In cases of 1-to-many, we take the first category. 
} 
Figure~\ref{figure:term_categories} shows the distribution of topic categories in \strategyqa{}. The distribution shows \strategyqa{} is very diverse, with the top two categories (``human'' and ``taxon'', i.e. a group of organisms) covering only a quarter of the data, and a total of 609 topic categories.

We further compare the diversity of \strategyqa{} to \textsc{HotpotQA}, a multi-hop QA dataset over Wikipedia paragraphs. To this end, we sample 739 pairs of evidence paragraphs associated with a single question in both datasets, and map the pair of paragraphs to a pair of Wikipedia categories using the ``instance of'' property.
We find that there are 571 unique category pairs in \strategyqa{}, but only 356 unique category pairs in \textsc{HotpotQA}.
Moreover, the top two category pairs in both of the datasets (``human-human'', ``taxon-taxon'') constitute 8\% and 27\% of the cases in \strategyqa{} and \textsc{HotpotQA}, respectively. This demonstrates the creativity and breadth of category combinations in \strategyqa{}.




\subsection{Human Performance}
\label{subsec:human_performance}

To see how well humans answer strategy questions, 
we sample a subset of 100 questions from \strategyqa{} and have experts (authors) answer questions, given access to Wikipedia articles and an option to reveal the decomposition for every question. In addition, we ask them to provide a short explanation for the answer, the number of searches they conducted to derive the answer, and to indicate whether they have used the decomposition. 
We expect humans to excel at coming up with strategies for answering questions. Yet, humans are not necessarily an upper bound because finding the relevant paragraphs is difficult and could potentially be performed better by machines.

Table~\ref{table:human_performance} summarizes the results. Overall, humans infer the required strategy and answer the questions with high accuracy. 
Moreover, the low number of searches shows that humans leverage background knowledge, as they can answer some of the intermediate steps without search.  
An error analysis shows that the main reason for failure (10\%) is difficulty to find evidence, and the rest of the cases (3\%) are due to ambiguity in the question that could lead to the opposite answer.

\begin{table}
    \centering
    \footnotesize
    \begin{tabular}{l|c}
         Answer accuracy & 87\%  \\
         Strategy match & 86\% \\
         Decomposition usage & 14\% \\
         Average \# searches & 1.25
    \end{tabular}
    \caption{Human performance in answering questions. Strategy match is computed by comparing the explanation provided by the expert with the decomposition. Decomposition usage and the number of searches are computed based on information provided by the expert.}
    \label{table:human_performance}
\end{table}

\section{Experimental Evaluation}
\label{sec:experiments}

In this section, we conduct experiments to answer the following questions: a) How well do pre-trained language models (LMs) answer strategy questions? b) Is retrieval of relevant context helpful? and c) Are decompositions useful for answering questions that require implicit knowledge?

\subsection{Baseline Models}
\label{subsec:baselines}


Answering strategy questions requires external knowledge that cannot be obtained by training on \strategyqa{} alone. Therefore, our models and online solvers (\S\ref{subsection:cqw}) are based on pre-trained LMs, fine-tuned on auxiliary datasets that require reasoning. Specifically, in all models we fine-tune \textsc{RoBERTa} \cite{liu2019roberta} on a subset of:
\begin{itemize}[leftmargin=*,topsep=5pt,itemsep=0pt,parsep=0pt]
    \item \textsc{BoolQ} \cite{clark2019boolq}: A dataset for boolean question answering.
    \item \textsc{MNLI} \cite{williams2018broad}: A large natural language inference (NLI) dataset. The task is to predict if a textual premise entails, contradicts or is neutral with respect to the hypothesis.
    \item \textsc{Twenty Questions} (\textsc{20Q}): A collection of 50K short commonsense boolean questions.\footnote{\url{https://github.com/allenai/twentyquestions}}
    \item \textsc{DROP} \cite{dua2019drop}: A large dataset for numerical reasoning over paragraphs.
\end{itemize}

Models are trained in two configurations:
\begin{itemize}[leftmargin=*,topsep=0pt,itemsep=0pt,parsep=0pt]
    \item \textbf{No context} 
    : The model is fed with the question only, and outputs a binary prediction using the special \texttt{CLS} token.

    \item \textbf{With context} 
    : We use \textsc{BM25} \cite{robertson1995okapi} to retrieve context from our corpus, while removing stop words from all queries. We examine two retrieval methods: a) question-based retrieval: by using the question as a query and taking the top $k=10$ results, and b) decomposition-based retrieval: by initiating a separate query for each (gold or predicted) decomposition step and concatenating the top $k=10$ results of all steps (sorted by retrieval score).
    In both cases, the model is fed with the question concatenated to the retrieved context, truncated to
    512 tokens (the maximum input length of \textsc{RoBERTa}), and outputs a binary prediction.
\end{itemize}


\paragraph{Predicting decompositions}
We train a seq-to-seq model, termed  \textsc{BART$_{\text{decomp}}$}, that given a question, generates its decomposition token-by-token. Specifically, we fine-tune \textsc{BART} \cite{lewis2020bart} on \strategyqa{} decompositions.

\paragraph{Baseline models}
As our base model, we train a model as follows: 
We take a \textsc{RoBERTa} \cite{liu2019roberta} model and fine-tune it on \textsc{DROP}, \textsc{20Q} and \textsc{BoolQ} (in this order). The model is trained on \textsc{DROP} with multiple output heads, as in \citet{segal2020simple}, which are then replaced with a single Boolean output.\footnote{For brevity, exact details on model training and hyper-parameters will be released as part of our codebase.} We call this model \textsc{RoBERTa*}.

We use \textsc{RoBERTa*} and \textsc{RoBERTa} to train the following models on \strategyqa{}: without context (\textsc{RoBERTa*$_\varnothing$}), with question-based retrieval (\textsc{RoBERTa*$_\text{IR-Q}$}, \textsc{RoBERTa$_\text{IR-Q}$}), and with predicted decomposition-based retrieval (\textsc{RoBERTa*$_\text{IR-D}$}). 

We also present four oracle models: 
\begin{itemize}[leftmargin=*,topsep=0pt,itemsep=0pt,parsep=0pt]
    
    \item \textsc{RoBERTa*$_\text{ORA-P}$}: uses the gold paragraphs (no retrieval).
    
    \item \textsc{RoBERTa*$_\text{IR-ORA-D}$}: performs retrieval with the gold decomposition.
    
    \item \textsc{RoBERTa*}$^{\text{last-step}}_\text{ORA-P-D}$: exploits both the gold decomposition and the gold paragraphs. We fine-tune \textsc{RoBERTa} on \textsc{BoolQ} and \textsc{SQuAD} \cite{rajpurkar2016squad} to obtain a model that can answer single-step questions. We then run this model on \strategyqa{} to obtain answers for all decomposition sub-questions, and replace all placeholder references with the predicted answers. Last, we fine-tune \textsc{RoBERTa*} to answer the last decomposition step of \strategyqa{}, for which we have supervision. 
    
    \item \textsc{RoBERTa*}$^{\text{last-step-raw}}_\text{ORA-P-D}$: \textsc{RoBERTa*} that is fine-tuned to predict the answer from the gold paragraphs and the last step of the gold decomposition, \emph{without} replacing placeholder references.

\end{itemize}


\paragraph{Online solvers}
For the solvers integrated in the data collection process (\S\ref{subsection:cqw}), we use three no-context models and two question-based retrieval models. The solvers are listed in Table~\ref{table:baseline_models}.

\begin{table}[t]
    \centering
    \footnotesize
    \begin{tabular}{p{4.4cm}c}
      Model  & Solver group(s) \\ \hline
      \textsc{RoBERTa$_\varnothing$(20Q)}  &  \ckptzero{}, \fntd{} \\
      \textsc{RoBERTa$_\varnothing$(20Q+BoolQ)}  & \ckptzero{}, \fntd{} \\
      \textsc{RoBERTa$_\varnothing$(BoolQ)}  &  \ckptzero{}, \fntd{} \\
      \textsc{RoBERTa$_{\text{IR-Q}}$(BoolQ)}  & \ckptzero{} \\
      \textsc{RoBERTa$_{\text{IR-Q}}$(MNLI+BoolQ)}  & \ckptzero{} 
    \end{tabular}
    \caption{QA models used as online solvers during data collection (\S\ref{subsection:cqw}).
    Each model was fine-tuned on the datasets mentioned in its name.
    }
    \label{table:baseline_models}
\end{table}


\subsection{Results}
\label{subsection:experiment_results}

\begin{table}[t]
    \centering
    \footnotesize
    \begin{tabular}{l|cc}
      Model & Accuracy & Recall@10 \\ \toprule
      \textsc{Majority} & 53.9 & - \\
      \textsc{RoBERTa*$_\varnothing$} & 63.6 $\pm$ 1.3 & - \\
      \textsc{RoBERTa$_{\text{IR-Q}}$} & 53.6 $\pm$ 1.0 & 0.174 \\
      \textsc{RoBERTa*$_{\text{IR-Q}}$} & 63.6 $\pm$ 1.0 & 0.174  \\
      \textsc{RoBERTa*$_{\text{IR-D}}$} & 61.7 $\pm$ 2.2 & 0.195 \\  \hline \hline
      \textsc{RoBERTa*$_{\text{IR-ORA-D}}$} & 62.0 $\pm$ 1.3 & 0.282 \\
      \textsc{RoBERTa*$_{\text{ORA-P}}$} & 70.7 $\pm$ 0.6 & - \\
       \textsc{RoBERTa*}$^{\text{last-step-raw}}_\text{ORA-P-D}$ & 65.2 $\pm$ 1.4 & -\\
       \textsc{RoBERTa*}$^{\text{last-step}}_\text{ORA-P-D}$ & 72.0 $\pm$ 1.0 & -\\
    \end{tabular}
    \caption{QA accuracy (with standard deviation across 7 experiments), and retrieval performance, measured by Recall@10, of baseline models on the test set.}
    \label{table:baseline_results}
\end{table}

\paragraph{Strategy QA performance}
Table~\ref{table:baseline_results} summarizes the results of all models (\S\ref{subsec:baselines}). \textsc{RoBERTa*$_{\text{IR-Q}}$} substantially outperforms \textsc{RoBERTa$_{\text{IR-Q}}$}, indicating that fine-tuning on related auxiliary datasets before \strategyqa{} is crucial. Hence, we focus on \textsc{RoBERTa*} for all other results and analysis.

Strategy questions pose a combined challenge of retrieving the relevant context, and deriving the answer based on that context. Training without context shows a large accuracy gain of $53.9 \rightarrow 63.6$ over the majority baseline. This is far from human performance, but shows that some questions can be answered by a large LM fine-tuned on related datasets without retrieval.
On the other end, training with \emph{gold} paragraphs raises performance to $70.7$. This shows that high-quality retrieval lets the model effectively reason over the given paragraphs. Last, using both gold decompositions and retrieval further increases performance to $72.0$, showing the utility of decompositions. 


Focusing on retrieval-based methods, we observe that question-based retrieval reaches an accuracy of $63.6$ and retrieval with gold decompositions results in an accuracy of $62.0$. This shows that the quality of retrieval even with gold decompositions is not high enough to improve the $63.6$ accuracy obtained by \textsc{RoBERTA*$_\varnothing$}, a model that uses no context. Retrieval with predicted decompositions results in an even lower accuracy of $61.7$. We also analyze predicted decompositions below.




\paragraph{Retrieval evaluation}
A question decomposition describes the reasoning steps for answering the question. Therefore, using the decomposition for retrieval may help obtain the relevant context and improve performance. 
To test this, we directly compare performance of question- and decomposition-based retrieval with respect to the annotated gold paragraphs. 
We compute Recall@10, i.e., the fraction of the gold paragraphs retrieved in the top-10 results of each method. Since there are 3 annotations per question, we compute Recall@10 for each annotation and take the maximum as the final score. For a fair comparison, in decomposition-based retrieval, we use the top-10 results across \emph{all} steps.


Results (Table~\ref{table:baseline_results}) show that retrieval performance is low, partially explaining why retrieval models do not improve performance compared to \textsc{RoBERTa*$_\varnothing$},
and demonstrating the retrieval challenge in our setup. Gold decomposition-based retrieval substantially outperforms question-based retrieval, showing that using the decomposition for retrieval is a promising direction for answering multi-step questions. Still, predicted decomposition-based retrieval does not improve retrieval compared to question-based retrieval, showing better decomposition models are needed.


To understand the low retrieval scores, we analyzed the query results of 50 random decomposition steps. Most failure cases are due to the shallow pattern matching done by BM25, e.g., failure to match synonyms. This shows that indeed there is little word overlap between decomposition steps and the evidence, as intended by our pipeline design. In other examples, either a key question entity was missing because it was represented by a reference token, or the decomposition step had complex language, leading to failed retrieval.
This analysis suggests that 
advances in neural retrieval might be beneficial for \strategyqa{}.

\paragraph{Human retrieval performance}
To quantify human performance in finding gold paragraphs, we ask experts to find evidence paragraphs for 100 random questions. For half of the questions we also provide decomposition. 
We observe average Recall@10 of $0.586$ and $0.513$ with and without the decomposition, respectively. This shows that humans significantly outperform our IR baselines. However, humans are still far from covering the gold paragraphs, since there are multiple valid evidence paragraphs (\S\ref{subsec:data_quality}), and retrieval can be difficult even for humans.
Lastly, using decompositions improves human retrieval, showing decompositions indeed are useful for finding evidence.

\paragraph{Predicted decompositions}
Analysis shows that \textsc{BART$_{\text{DECOMP}}$}'s decompositions are grammatical and well-structured. Interestingly, the model generates strategies, but often applies them to questions incorrectly.
E.g., the question \nl{Can a lifeboat rescue people in the Hooke Sea?} is decomposed to \nl{1) What is the maximum depth of the Hooke Sea? 2) How deep can a lifeboat dive? 3) Is \#2 greater than or equal to \#1?}. While the decomposition is  well-structured, it uses a wrong strategy (lifeboats do not dive).

\section{Related Work}
\label{sec:related_work}

Prior work has typically let annotators write questions based on an entire context \cite{khot2020qasc, yang2018hotpotqa, dua2019drop, mihaylovetal2018, khashabi2018looking}. 
In this work, we prime annotators with minimal information (few tokens) and let them use their imagination and own wording to create questions. 
A related priming method was recently proposed by \citet{clark2020tydi}, who used the first 100 characters of a Wikipedia page. 

Among multi-hop reasoning datasets, our dataset stands out in that it requires \emph{implicit} decompositions.
Two recent datasets \cite{khot2020qasc, mihaylovetal2018} have considered questions requiring implicit facts. However, they are limited to specific domain strategies, while in our work we seek diversity in this aspect.

Most multi-hop reasoning datasets do not fully annotate question decomposition \cite{yang2018hotpotqa, khot2020qasc, mihaylovetal2018}. 
This issue has prompted recent work to create question decompositions for existing datasets~\cite{wolfson2020break}, and to train models that generate question decompositions \cite{perez2020unsupervised, khot2020text, min2019multi}.
In this work, we annotate question decompositions as part of the data collection. 


\section{Conclusion}
\label{sec:conclusion}

We present \strategyqa{}, the first dataset of \emph{implicit} multi-step questions requiring a wide-range of reasoning skills.
To build \strategyqa{}, we introduced a novel annotation pipeline for eliciting creative questions that use simple language, but cover a challenging range of diverse strategies.
Questions in \strategyqa{} are annotated with decomposition into reasoning steps and evidence paragraphs, to guide the ongoing research towards addressing implicit multi-hop reasoning. 


\section*{Acknowledgement}
We thank Tomer Wolfson for helpful feedback and the REVIZ team at Allen Institute for AI, particularly Michal Guerquin and Sam Skjonsberg.
This research was supported in part by the Yandex Initiative for Machine Learning, and the European Research Council (ERC) under the European Union Horizons 2020 research and innovation programme (grant ERC DELPHI 802800). Dan Roth is partly supported by ONR contract N00014-19-1-2620 and DARPA contract FA8750-19-2-1004, under the Kairos program.  
This work was completed in partial fulfillment for the Ph.D degree of Mor Geva.

\bibliography{all}
\bibliographystyle{acl_natbib}

\end{document}